\documentclass[sigconf]{acmart}

\usepackage{mathtools}
\usepackage{balance}
\usepackage{booktabs}
\usepackage{tikz}
\usepackage{xcolor}
\usepackage{cancel}
\usepackage{soul}
\usepackage{eucal}
\usepackage{url}
\usepackage{float}
\usepackage{amsmath,amsfonts}
\usepackage{amsthm, mathtools}
\usepackage{booktabs}
\usepackage{algorithm}
\usepackage{algpseudocode}
\usepackage{multirow, multicol}
\usepackage{subfig}

\newtheorem{remark}{Remark}
\AtBeginDocument{%
  }

\copyrightyear{2025}
\acmYear{2025}
\setcopyright{acmlicensed}\acmConference[CIKM '25]{Proceedings of the 34th ACM International Conference on Information and Knowledge Management}{November 10--14, 2025}{Seoul, Republic of Korea}
\acmBooktitle{Proceedings of the 34th ACM International Conference on Information and Knowledge Management (CIKM '25), November 10--14, 2025, Seoul, Republic of Korea}
\acmDOI{10.1145/3746252.3761288}
\acmISBN{979-8-4007-2040-6/2025/11}
\begin{document}

\title{Efficiency Boost in Decentralized Optimization: Reimagining Neighborhood Aggregation with Minimal Overhead}


\author{Durgesh Kalwar}
\affiliation{%
  \institution{Arizona State University}
  \city{Tempe}
  \state{Arizona}
  \country{USA}}
\email{dkalwar@asu.edu}

\author{Mayank Baranwal}
\affiliation{%
  \institution{TCS Research}
  \city{Mumbai}
  \state{Maharahstra}
  \country{India}}
\email{baranwal.mayank@tcs.com}

\author{Harshad Khadilkar}
\affiliation{%
 \institution{Indian Institute of Technology, Bombay}
 \city{Mumbai}
 \state{Maharahstra}
 \country{India}}
\email{harshadk@iitb.ac.in}






\begin{abstract}
  In today's data-sensitive landscape, distributed learning emerges as a vital tool, not only fortifying privacy measures but also streamlining computational operations. This becomes especially crucial within fully decentralized infrastructures where local processing is imperative due to the absence of centralized aggregation. Here, we introduce \texttt{DYNAWEIGHT}, a novel framework to information aggregation in multi-agent networks. \texttt{DYNAWEIGHT} offers substantial acceleration in decentralized learning with minimal additional communication and memory overhead. Unlike traditional static weight assignments, such as Metropolis weights, \texttt{DYNAWEIGHT} dynamically allocates weights to neighboring servers based on their relative losses on local datasets. Consequently, it favors servers possessing diverse information, particularly in scenarios of substantial data heterogeneity. Our experiments on various datasets MNIST, CIFAR10, and CIFAR100 incorporating  various server counts and graph topologies, demonstrate notable enhancements in training speeds. Notably, \texttt{DYNAWEIGHT} functions as an aggregation scheme compatible with any underlying server-level optimization algorithm, underscoring its versatility and potential for widespread integration.
\end{abstract}

\ccsdesc[500]{Computing methodologies~Distributed algorithms}
\ccsdesc[300]{Computing methodologies~Multi-agent systems}
\ccsdesc[300]{Computing methodologies~Cooperation and coordination}

\keywords{Decentralized Optimization, Consensus, Dynamic Weighting, Distributed Learning}


\maketitle
\section{Introduction}\label{sec:Intro}
Distributed and decentralized optimization\footnote{In the literature, ``distributed" and ``decentralized" are frequently used interchangeably. While they differ slightly, both terms denote a setup lacking a central server or compute. Each server processes private data locally, with information exchanged between neighboring servers.} techniques have emerged as pivotal strategies in tackling large-scale computational challenges across diverse domains~\cite{li2019communication, elgabli2020communication, koloskova2020unified}. In distributed optimization, the computational task is divided among multiple processing units or nodes, each contributing to the overall solution. This approach offers advantages such as scalability, fault tolerance, and parallelism. Decentralized optimization takes this concept further by removing the need for a central coordinator, allowing individual nodes to collaboratively reach a solution through local interactions.

The importance of decentralized learning is underscored by their applicability in various fields. In machine learning and artificial intelligence, these techniques enable the training of complex models on massive datasets distributed across different locations, while preserving data privacy and security~\cite{xin2020decentralized}. Decentralized optimization finds applications in decentralized control systems~\cite{inalhan2002decentralized}, sensor networks~\cite{rabbat2004distributed}, and multi-agent systems~\cite{baranwal2020robust}, where local decision-making leads to emergent global behavior. Moreover, in fields like logistics and transportation, these approaches facilitate real-time decision-making in dynamic environments, optimizing resource allocation and routing.

However, despite their immense potential, distributed and decentralized optimization methods come with several challenges. One primary concern is communication overhead, as exchanging information between nodes can lead to bottlenecks and increased latency~\cite{ballotta2023can}. Ensuring convergence to a global optimum in decentralized settings without a central coordinator poses another challenge, requiring sophisticated algorithms that balance exploration and exploitation~\cite{taylor2011distributed}. Moreover, maintaining consistency and synchronization across distributed systems in the presence of failures or network delays is a non-trivial task~\cite{hsieh2022multi}. Additionally, privacy and security concerns arise when dealing with sensitive data distributed across multiple nodes, necessitating robust encryption and access control mechanisms~\cite{zhang2018enabling}.

Addressing these challenges requires a multidisciplinary approach, integrating techniques from optimization theory, distributed systems, machine learning, and cryptography. Advances in communication protocols, consensus algorithms, and optimization methods tailored for distributed and decentralized environments are crucial. Furthermore, developing robust frameworks for evaluating the performance and scalability of these techniques across diverse applications is essential for their widespread adoption.

In this paper, we introduce \texttt{DYNAWEIGHT}, an adaptive weighting framework designed to address data heterogeneity across servers in decentralized learning. Unlike static weighting schemes that rely solely on network connectivity for server weighting during information aggregation, \texttt{DYNAWEIGHT} utilizes server performance on the datasets of neighboring servers (without explicitly sharing the data). We empirically demonstrate that \texttt{DYNAWEIGHT} achieves faster convergence compared to traditional non-adaptive weighting schemes while remaining computationally efficient. Moreover, \texttt{DYNAWEIGHT} is a versatile weighting scheme that can be integrated with any suitable optimization algorithm.


\section{Distributed Learning in Multi-Agent Systems}\label{sec:DLMAS}
We consider a network comprising $N$ interconnected servers (agents) over a communication graph $\mathcal{G}(V,\mathcal{E})$ collaboratively training a deep neural network in a decentralized fashion. Each server $i$ possesses its private dataset $\mathcal{D}_i\coloneqq (\mathbf{x}_i^{q},y_i^{q})_{q=1}^{n_i}$ comprising $n_i$ samples and maintains its individual network copy with parameters $\theta_i\in\mathbb{R}^d$. While a server $i$ can exchange parameter vectors with its neighbors $\mathcal{N}_i$, the confidentiality of local data is preserved. Homogeneity of data distribution across servers is not mandated. It's crucial to note that this setup fundamentally differs from conventional federated learning (FL), which resembles a centralized framework where a central server communicates with a fraction of edge devices for information exchange. The collective objective is to minimize the team objective function:
\begin{align}\label{eq:CumLoss}
    F(\theta) \coloneqq \sum\limits_{i=1}^N\frac{1}{n_i}\sum\limits_{q=1}^{n_i}\ell\left(\texttt{NN}_{\theta}(\mathbf{x}_i^q),y_i^q\right),
\end{align}
where $\texttt{NN}_{\theta}(\mathbf{x}_i^q)$ denotes the output of the neural network of the $i^\text{th}$ server given input $\mathbf{x}_i^q$, and $\ell(\cdot)$ represents any sample-level loss function. However, \eqref{eq:CumLoss} depicts a centralized framework for minimizing the cumulative loss function, assuming that the network parameters $\theta$ are shared across servers. A distributed version of the above problem can instead be expressed as:
\begin{align}\label{eq:DistLoss}
     \min\limits_{\{\theta_1,\theta_2,\dots,\theta_N\}}&\sum\limits_{i=1}^N\frac{1}{n_i}\sum\limits_{q=1}^{n_i}\ell\left(\texttt{NN}_{\theta_i}(\mathbf{x}_i^q),y_i^q\right) \nonumber \\
     \text{s.t.} \quad &\theta_1 = \theta_2 = \dots = \theta_N.
\end{align}
However, in the absence of a centralized server, servers aim to execute a consensus algorithm to ensure convergence of their parameters to a common vector. A typical decentralized optimization algorithm executes following steps at $k^\text{th}$ iteration of each server:
\begin{align*}
     \theta_i^{k+\frac{1}{2}} &\gets \theta_i^{k} - \eta \frac{1}{n_i}\sum\limits_{q=1}^{n_i}\nabla_{\theta_i}\ell\left(\texttt{NN}_{\theta_i}(\mathbf{x}_i^q),y_i^q\right), \tag{Gradient step}  \\
     \theta_i^{k+1} &\gets \sum\limits_{j\in\mathcal{N}_i}w_{ij}\theta_j^{k+\frac{1}{2}}, \tag{Consensus}
\end{align*}
where $\eta>0$ denotes the step-size, while weights $w_{ij}\geq 0$, with $\sum_{j\in\mathcal{N}i}w{ij}=1$, signify the relative importance of the $i^\text{th}$ server and its neighbors in updating the parameter vector $\theta_i$. It's important to recall that during the gradient step, each server computes the gradient of the loss function solely based on its local (private) data and executes a gradient descent. Furthermore, in practical implementations, this update process can be further accelerated by utilizing momentum-based optimization algorithms, such as AdaGrad~\cite{chen2023convergence} or Adam~\cite{nazari2022dadam}.

The next phase involves executing a consensus update among neighboring servers utilizing weights $w_{ij}$. There exist various methods for selecting these consensus weights. The prevalent approach involves employing equal weighting across all servers, termed as the simple weighting scheme. Alternatively, another viable option is the utilization of Metropolis weights, closely associated with the Metropolis-Hastings algorithm. The Metropolis weights are given as:
\begin{align}\label{eq:Metro}
    w_{ij} = \left\{\begin{array}{cl}
        \frac{1}{1+\max{(d_i, d_j)}}, &  \text{if} \ j\in\mathcal{N}_i, i\neq j\\
         1 - \sum_{l\in\mathcal{N}_i}w_{il}, & \text{if} \ i=j \\
         0, & \text{else}
    \end{array}\right.,
\end{align}
where $d_i$ represents the degree of $i^\text{th}$-server. It is worth noting that Metropolis weights can be derived through local information exchange, obviating the need for servers to possess knowledge of the entire communication topology.

\textbf{Heterogeneous data}: A notable challenge with static (fixed) weighting schemes lies in their inability to accommodate data heterogeneity across servers. To illustrate, consider an extreme scenario where we aim to train a DNN with $K$ classes. In this scenario, the first $(N-1)$ servers exhibit uniformly distributed data across $(K-1)$ classes, while the last server exclusively accesses data points from the $K^\text{th}$ class. Following the gradient step, the parameter vectors of the first $N-1$ servers align closely, whereas the parameters of the $N^\text{th}$ server diverge significantly. However, if servers employ, for instance, an equal weighting scheme, the information from the last server may be disregarded during the consensus step. Consequently, the remaining servers have limited opportunity to effectively learn to predict the $K^\text{th}$ class with reasonable accuracy. This limitation underscores the necessity for an \emph{adaptive} weighting scheme, one that dynamically adjusts weights based on gradient updates and alignment with other servers.



\section{Related Work}\label{sec:Related}
\textbf{Federated learning (FL)}: Standard practice for training neural networks involves collecting all data on a central server for model training, which offers simplicity and potentially higher accuracy due to access to comprehensive datasets. However, this method poses significant privacy risks and can lead to high communication costs~\cite{abdulrahman2020survey}. Federated Learning (FL), on the other hand, enables decentralized training by keeping data on local devices and only sharing model updates~\cite{mcmahan2017communication}. This approach enhances privacy~\cite{wei2020federated} and reduces data transfer costs but faces challenges such as data heterogeneity~\cite{qu2022rethinking}, slower convergence~\cite{wu2022node}, and increased complexity in coordinating multiple devices. Additionally, FL's reliance on a central server introduces a single point of failure and limits scalability due to significant communication bandwidth requirements. To overcome these challenges, decentralized learning is preferred~\cite{hegedHus2021decentralized}, as it distributes both data and computational tasks across multiple nodes, eliminating the need for a central server.

\noindent \textbf{Decentralized learning} involves two primary steps: (a) the Gradient step, where each server computes gradients on its private data and performs a gradient descent (or its momentum variant) on its parameter values, and (b) the Consensus or Gossip step, where servers aggregate parameter values from neighboring servers. Enhancing the speed of decentralized learning relies on three fundamental concepts: crafting efficient optimization algorithms for streamlined gradient updates, advancing faster consensus algorithms, and devising optimal communication topologies~\cite{vogels2022beyond, palmieri2023effect, jiang2023joint, song2022communication} to balance the trade-off between communication and computation efficiency.

The literature on optimization algorithms spans from simple gradient-based updates~\cite{yuan2016convergence} to momentum-based approaches like AdaGrad~\cite{chen2023convergence} and Adam~\cite{nazari2022dadam}. Similarly, gossip algorithms typically employ standard weighting schemes such as simple gossip, max-degree~\cite{kenyeres2022optimally}, and Metropolis weights~\cite{xiao2006distributed}. Recent studies suggest that static exponential topology~\cite{ying2021exponential, neglia2020decentralized} is efficient across commonly used topologies like ring, star, and line. However, in multi-agent networks, topology often emerges based on factors like geographical proximity, limiting control over it.

While selecting an efficient optimization algorithm for the gradient step is feasible, further progress depends on faster consensus updates~\cite{liu2019distributed, giuseppi2022weighted}. Unfortunately, beyond basic weighting schemes, there have been limited advancements in accelerating the Gossip step. This limitation is especially critical in scenarios of extreme data heterogeneity, where simple gossip algorithms exhibit slow convergence.
\section{DYNAWEIGHT: Dynamic Neighborhood Weighting via Train Loss}\label{sec:DYNAWEIGHT}
\begin{figure*}[ht]
	\begin{center}
		\begin{tabular}{ccc}
			\includegraphics[width=0.63\columnwidth]{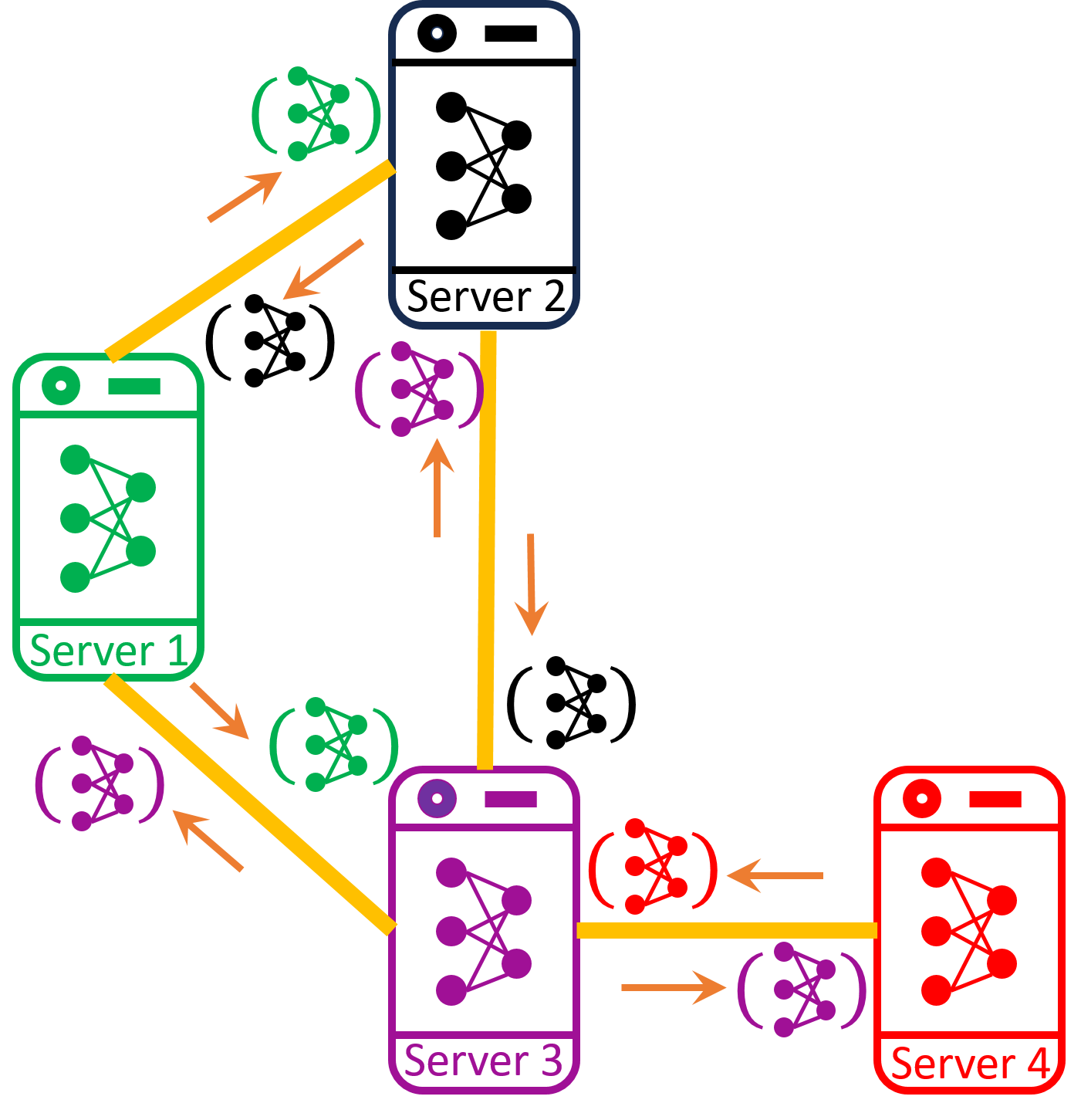} &
	       \includegraphics[width=0.63\columnwidth]{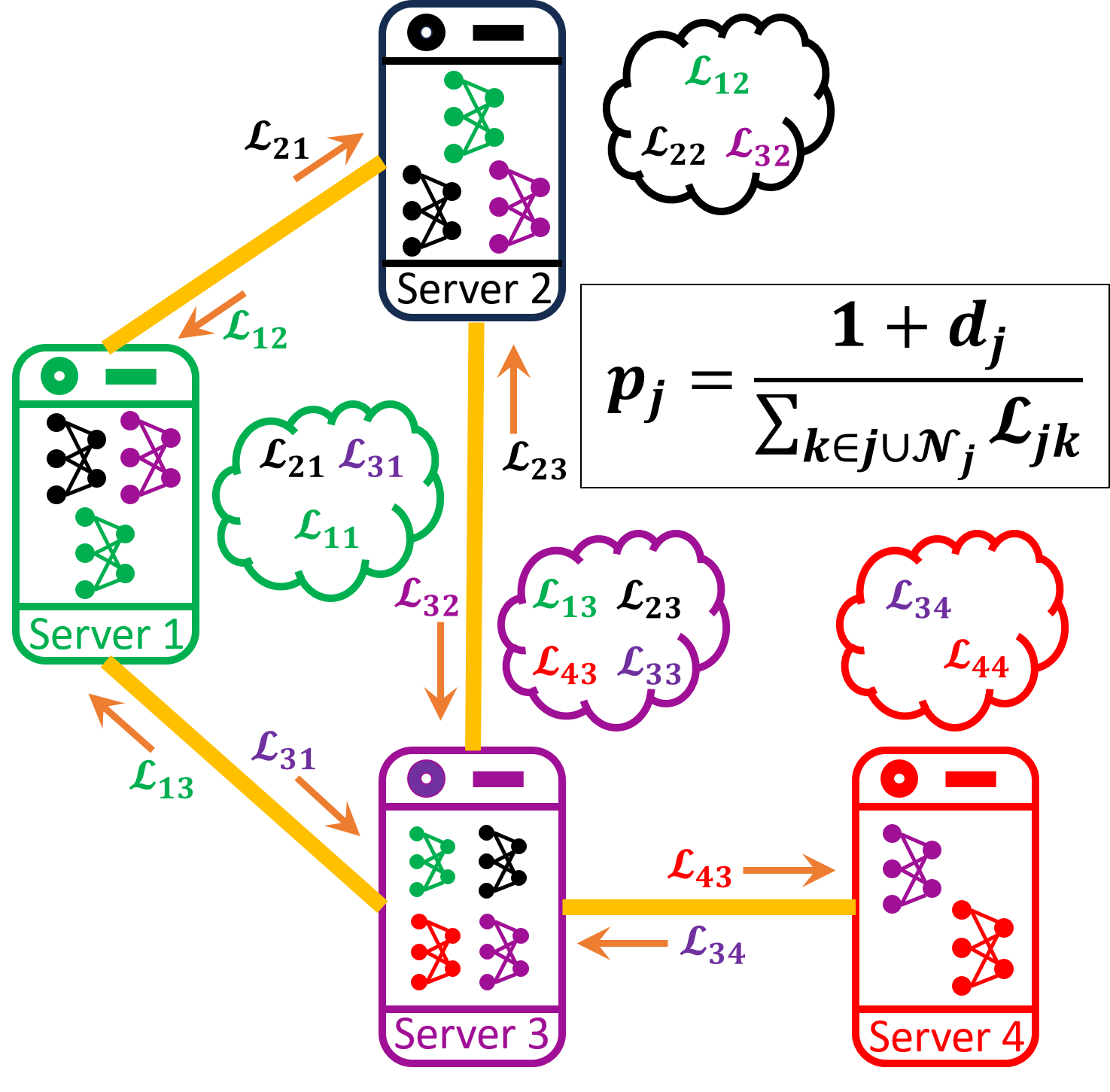}&
			\includegraphics[width=0.63\columnwidth]{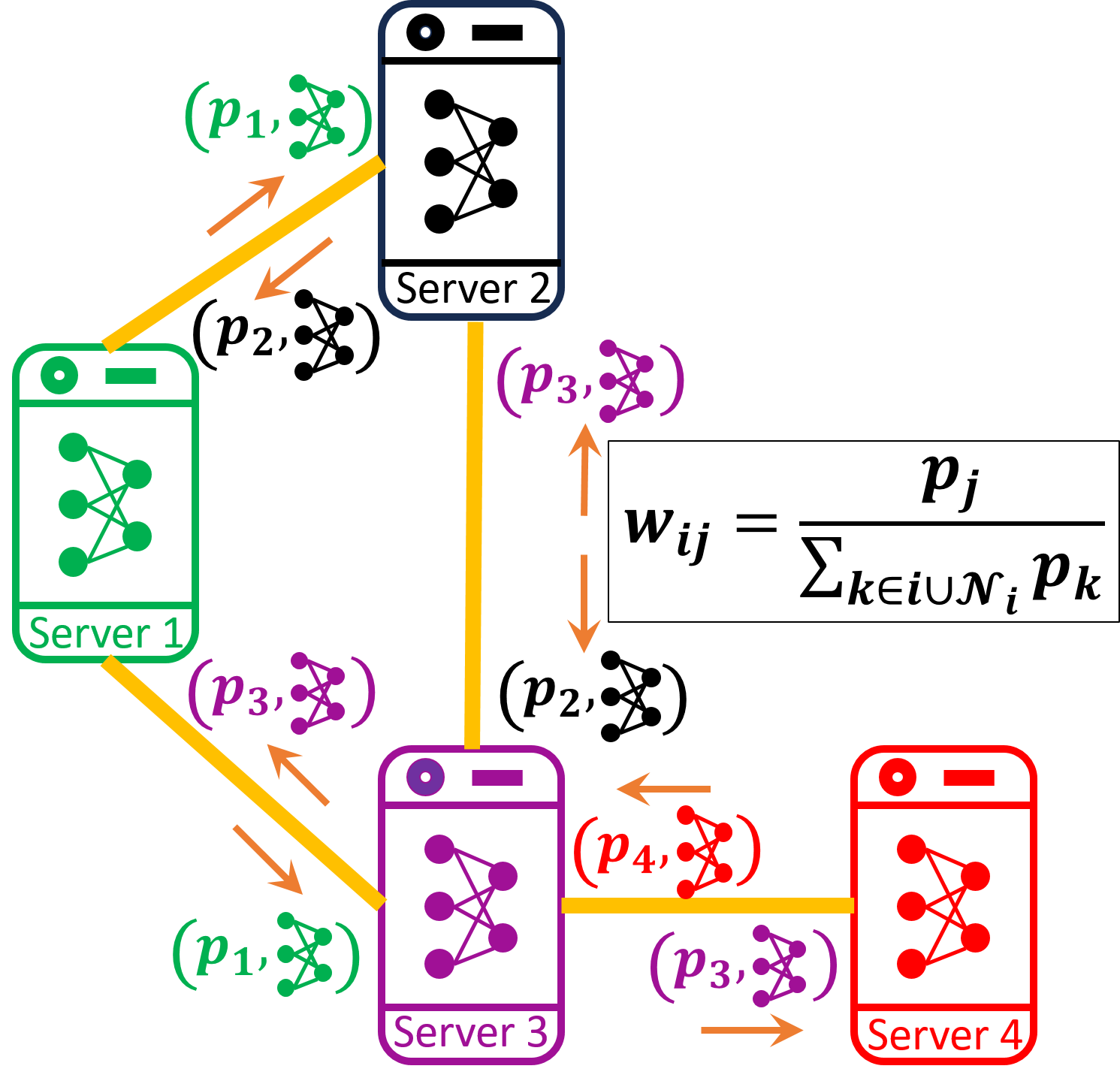}\cr 
            (a) & (b) & (c)
		\end{tabular}
        \end{center}
        \caption{Adaptive weighting via \texttt{DYNAWEIGHT}. (a) Each server broadcasts its locally updated parameters $\{\theta_i^{k+\frac{1}{2}}\}$ to its neighbors. (b) Upon receiving parameters from neighbors, servers perform \texttt{ReadOut($\cdot$)} and evaluate loss on their private datasets using neighbors parameters. This allows each server to evaluate its importance $\{p_i^k\}$. (c) Along with $\{\theta_i^{k+\frac{1}{2}}\}$, the importance $\{p_i\}$ is shared to locally evaluate aggregation weights $\{w_{ij}^k\}$.}
        \label{fig:DYNAWEIGHT}
\end{figure*}

\textbf{Intuition behind dynamic weighting}: When data distribution across servers is similar, one would naturally anticipate similarity in their parameter vectors. However, if one server exhibits significantly different data distribution, neighboring servers' models evaluated on this dataset would yield higher loss values. Consequently, neighboring servers would need to adjust their parameters accordingly. As parameter vectors start aligning, the relative importance of servers should also adjust accordingly. Static weighting schemes, however, neglect data distribution and base weights solely on connectivity. This underscores the necessity of weighting neighboring servers during consensus updates based on their model evaluation statistics on a server's own private dataset.

In a typical distributed learning framework, servers exchange their network parameters solely with their neighboring servers. Expanding on this information exchange, we additionally utilize the shared parameters to assess the neighboring server's model performance on the dataset of the server with which parameters have been exchanged. For instance, let us consider servers $i$ and $j$ as neighbors. When server $j$ shares its updated parameters ($\theta_j^{k+\frac{1}{2}}$) with the $i^\text{th}$-server (and vice versa), apart from directly conducting consensus, the $i^\text{th}$ server also evaluates (\emph{without storing}) the model $\text{NN}_{\theta_j}$ on its private dataset $\mathcal{D}_i$ and computes the loss $\mathcal{L}_{ji}$. This procedure is independently executed by each server.

From the perspective of the $i^\text{th}$-server, if the losses $\mathcal{L}_{ii}$ and $\mathcal{L}_{ji}$ differ significantly, this indicates a mismatch in the underlying data distributions between the two servers. Specifically, if $\mathcal{L}_{ji} > \mathcal{L}_{ii}$, it suggests that server $j$'s model does not perform well on the $i^\text{th}$-server's data. On the other hand, from the perspective of the $j^\text{th}$-server, if its model consistently achieves lower loss values across all its neighboring servers on average, it indicates that server $j$'s model is well-trained and performs effectively across the diverse datasets of its neighbors. Consequently, neighboring servers should aim to align their parameters with those of server $j$. This alignment can be achieved by prioritizing server $j$ during the consensus step.

We formalize this intuition using a novel dynamic weighting framework, \texttt{DYNAWEIGHT}. \texttt{DYNAWEIGHT} adaptively weighs servers based on the performance of their models on the private datasets of neighboring servers, without ever accessing those datasets directly. The framework is described in Figure~\ref{fig:DYNAWEIGHT}. After the gradient step at $k^\text{th}$-iteration, each server updates their parameters as $\{\theta_i^{k+\frac{1}{2}}\}$. The consensus step comprises of three phases:

\noindent\textbf{1) Readout Phase}: In this phase, servers exchange their parameter vectors with neighboring servers. The received model parameters are then used by the servers in the subsequent evaluation phase.

\noindent\textbf{2) Evaluation Phase}: Each server evaluates the models of its neighboring servers on its own local dataset and records the loss values ${\mathcal{L}{ji}}$. Here, $\mathcal{L}{ji}$ represents the loss evaluated by the $i^\text{th}$ server on its dataset using the $j^\text{th}$ server's parameters. These loss values are then communicated back to the respective servers, meaning server $i$ sends the loss value $\mathcal{L}_{ji}$ to server $j$, and so forth. Note that the loss values are scalars, and exchanging them entails minimal communication cost. Upon receiving the respective loss values, each server computes its importance (centrality) $p_j$ given as:
\begin{align}\label{eq:pj}
    p_j = \frac{1+d_j}{\sum_{m\in j\cup\mathcal{N}_j}\mathcal{L}_{jm}},
\end{align}
where $d_j$ represents the degree of the $j^\text{th}$-server. Mathematically, $p_j$ is interpreted as the inverse of the average loss values of the $j^\text{th}$-server on its own data and the data of its neighbors. We consider the inverse of the average loss because servers with smaller average loss are likely to perform better on the datasets of other servers as well. Therefore, other servers should aim to align their parameters closely with those having larger centrality.

\noindent \textbf{3) Gossip Phase}: In the final phase of the consensus step, servers execute a weighted gossip algorithm. The weights are based on the centrality of different servers computed in the previous phase. This phase begins with servers sending their centrality values to their neighboring servers. As before, centrality is a scalar quantity and entails minimal communication requirements. This exchange is combined with the previously updated parameter values ${\theta_i^{k+\frac{1}{2}}}$ shared during the readout phase. Based on the centrality values, each server $i$ computes the aggregation weights $w_{ij}$ as follows:
\begin{align}\label{eq:wij}
    w_{ij} = \frac{p_j}{\sum_{k\in i\cup\mathcal{N}_i}p_k}.
\end{align}
The overall algorithm is described in Algorithm~\ref{alg:DYNAWEIGHT}.

\begin{algorithm}[ht]
\caption{Distributed Learning (\texttt{DYNAWEIGHT})}
\label{alg:DYNAWEIGHT}
\begin{algorithmic}[1]
    \Require{$N$ servers, sub datasets $(\mathcal{D}_{1},\mathcal{D}_{2},\dots,\mathcal{D}_{N})$, communication topology $\mathcal{G}(V,\mathcal{E})$, \# of epochs $K$, \# of consensus steps $C$}\\
    \text{Randomly initialize the model parameter $\theta_{i}^0$ for each server}
    \For{each epoch $k=0,1,\dots,K-1$}
        \For{each server $i=1,2,....,N$}
            \For{each mini batch $b$ from $\mathcal{D}_{i}$}
                \State $\theta_{i}^{k+\frac{1}{2}} \leftarrow \theta_{i}^{k} - \eta\frac{1}{n_i}\sum_{\mathcal{D}_i}\nabla_{\theta_i} \ell(b|\theta_{i}^{k})$ \Comment{Gradient step}
            \EndFor
        \EndFor
        \For{each server $i=1,2,....,N$}
            \State \text{Broadcast $\{\theta_{i}^{k+\frac{1}{2}}\}$ to neighbors} \Comment{Readout}
            \For{each server $j\in \mathcal{N}_i$}
                \State Evaluate $\{\mathcal{L}_{ji}\}$ on local dataset
                \State Broadcast back $\{\mathcal{L}_{ji}\}$ to server $j$
            \EndFor\Comment{Evaluation}
            \State Compute centrality $p_i \gets \frac{1+d_i}{\sum_{m\in i\cup\mathcal{N}_i}\mathcal{L}_{im}}$
            \For{each server $j\in \mathcal{N}_i$}
                \State $w_{ij} \gets \frac{p_j}{\sum_{q\in i\cup\mathcal{N}_ip_q}}$
            \EndFor
            \For{consensus step $c=1,2,....,C$}
                \State $\theta_i^k \gets \sum_{j\in\mathcal{N}_i}w_{ij}\theta_j^{k+\frac{1}{2}}$ \Comment{Gossip}
            \EndFor
        \EndFor
    \EndFor
\end{algorithmic}
\end{algorithm}

\begin{remark}
    In \texttt{DYNAWEIGHT}, servers also maintain a \emph{ghost} copy of their neural network architecture. This ghost copy is used to evaluate the loss values on their private datasets after receiving the parameters from neighboring servers. Additionally, the same shared network parameters $\theta_i^{k+\frac{1}{2}}$ are accessed twice: once during the readout phase and again during the gossip phase. Unlike standard non-adaptive weighting schemes, \texttt{DYNAWEIGHT} also requires servers to broadcast loss and centrality values, both of which are scalars. Consequently, the additional computational and memory overheads are minimal compared to static weighting schemes.
\end{remark}


\section{Experimental Results}\label{sec:Exp}
In this section, we present the evaluation of the proposed framework, \texttt{DYNAWEIGHT}, on the widely-used classification datasets: MNIST, CIFAR10 and CIFAR100. The experiments are conducted across various model architectures and under multiple graph topologies, namely ring, line, chordal, and static exponential, while varying the number of servers, $N = 8, 16,$ and $32$. To provide a comprehensive performance comparison, we benchmark \texttt{DYNAWEIGHT} against several baseline approaches:


\begin{figure*}[htbp]
    \centering
    \begin{minipage}{\textwidth}
        \centering
        \includegraphics[width=0.88\textwidth]{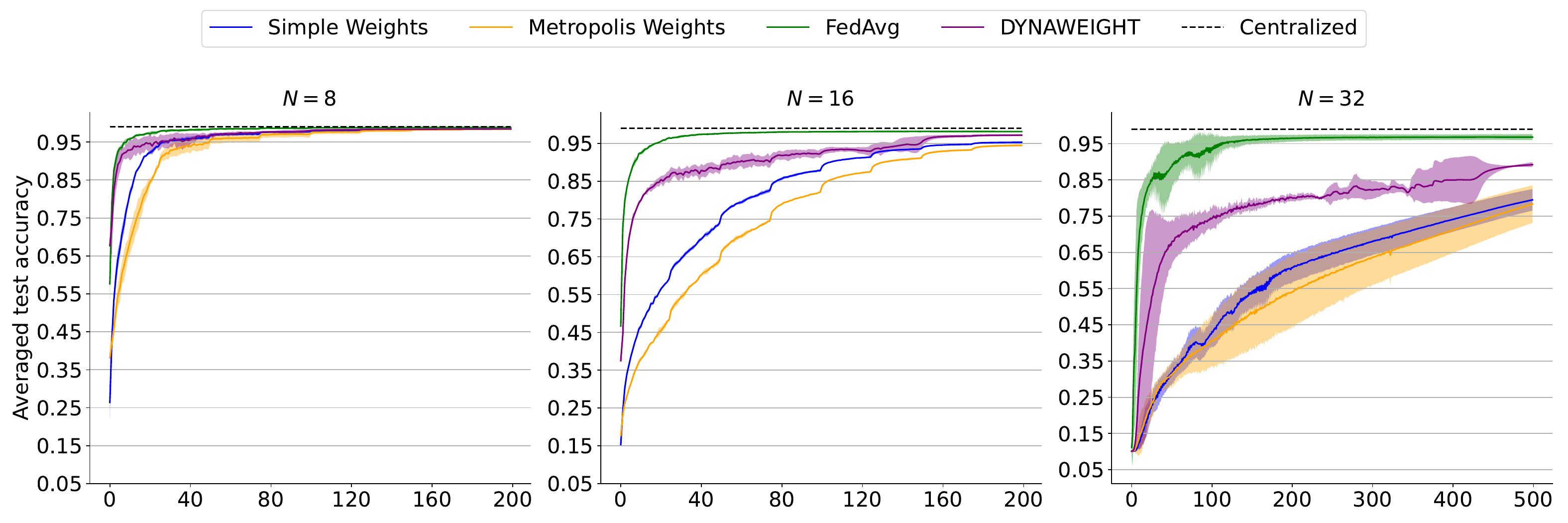}\\
        {\small (a) Ring Topology}
    \end{minipage}
    
    \begin{minipage}{\textwidth}
        \centering
        \includegraphics[width=0.88\textwidth]{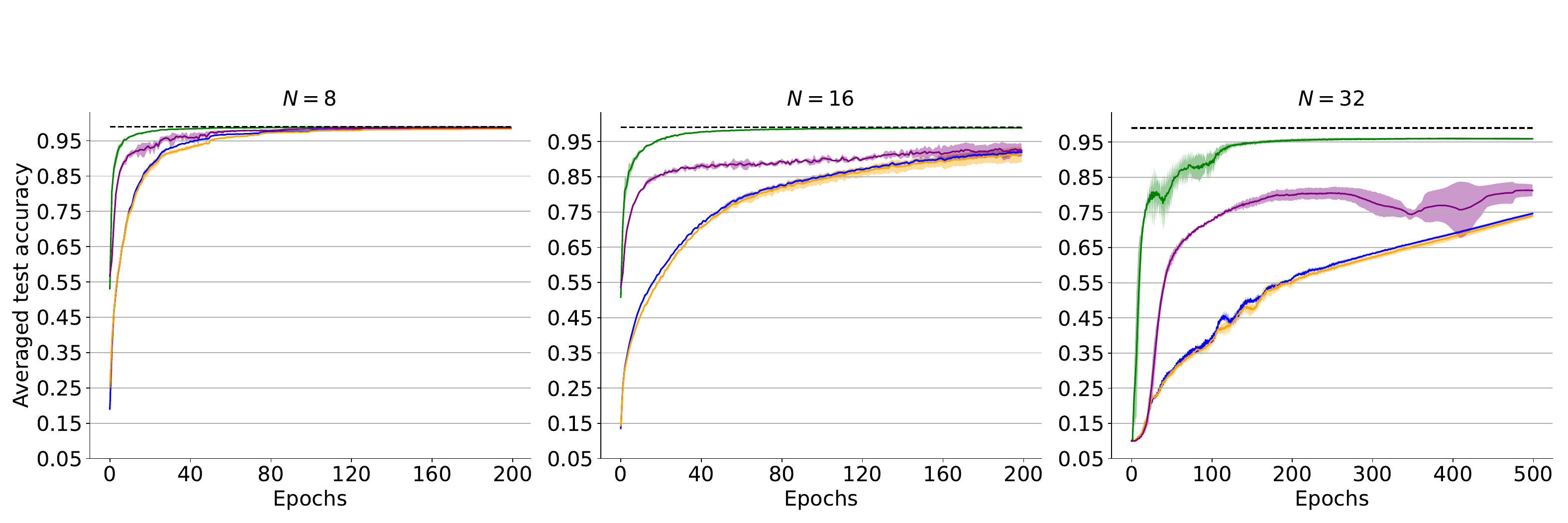}\\
        {\small (b) Line Topology}
    \end{minipage}
    
    \begin{minipage}{\textwidth}
        \centering
        \includegraphics[width=0.88\textwidth]{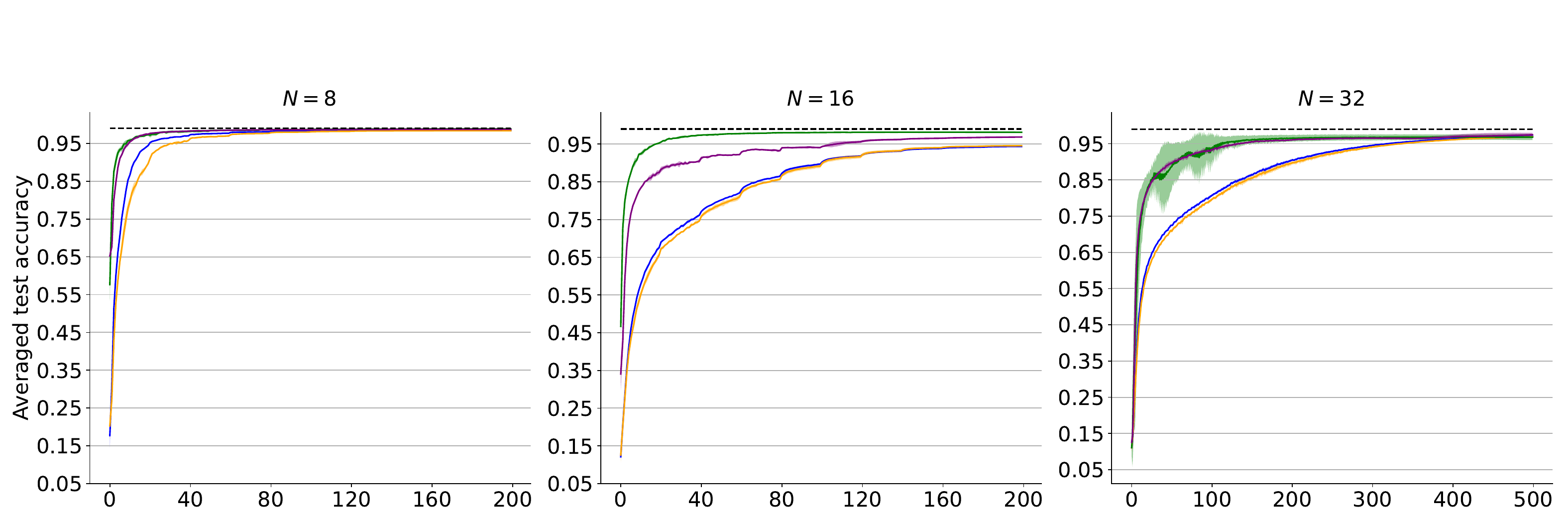}\\
        {\small (c) Chordal Topology}
    \end{minipage}
    
    \caption{Performance comparison of all approaches on the MNIST dataset with LeNet architecture for various graph topologies, evaluated across different server counts ($N = 8, 16,$ and $32$). The y-axis represents the average test accuracy, computed over $3$ random seeds and averaged across all servers. The shaded regions indicate the $95\%$ confidence interval.}
    \label{fig:MNIST_test_acc}
\end{figure*}


\begin{figure*}[htbp]
    \centering
    \begin{minipage}{\textwidth}
        \centering
        \includegraphics[width=0.88\textwidth]{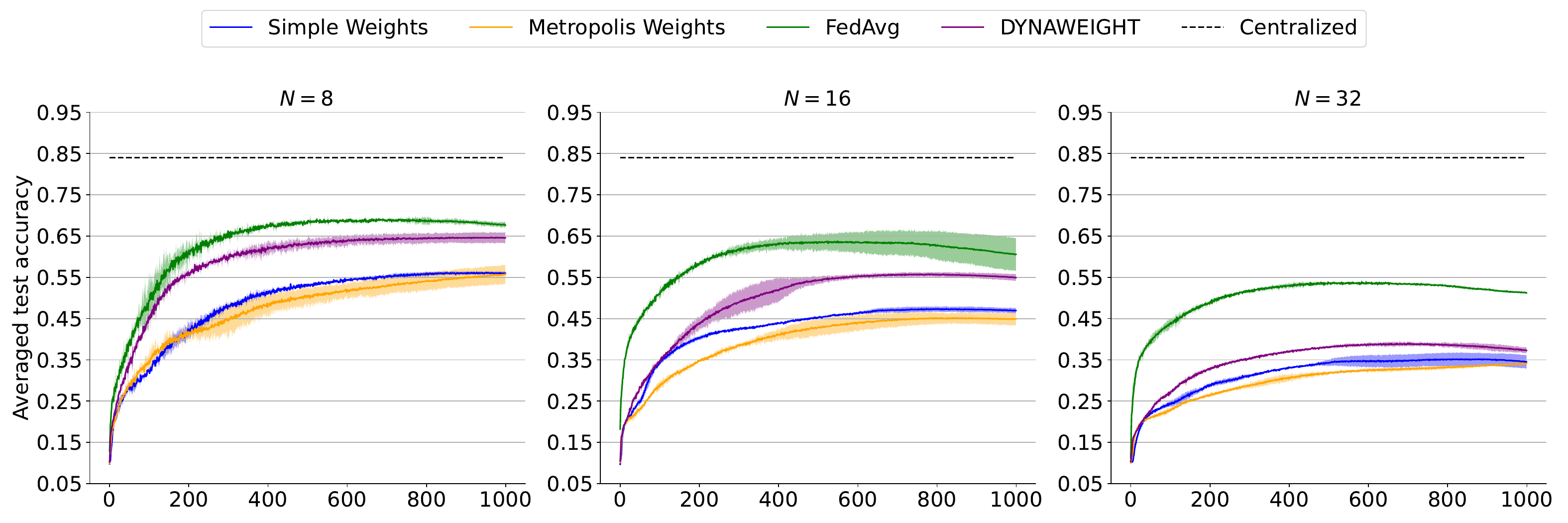}\\
        {\small (a) Ring Topology}
    \end{minipage}
    
    \begin{minipage}{\textwidth}
        \centering
        \includegraphics[width=0.88\textwidth]{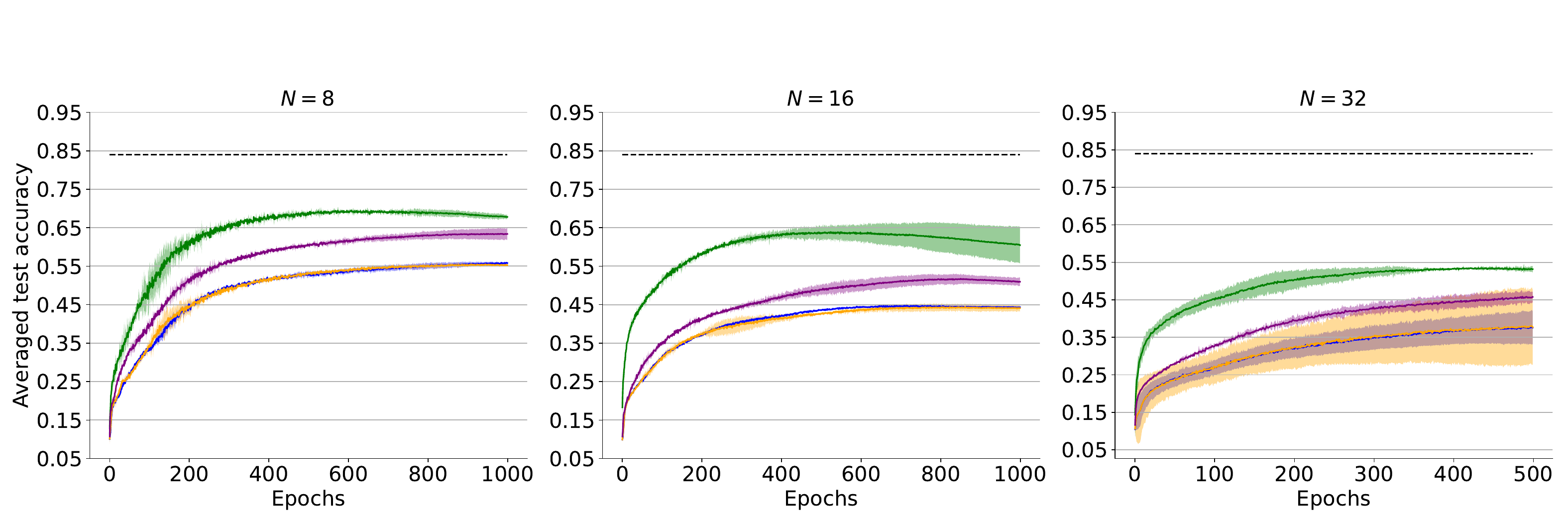}\\
        {\small (b) Line Topology}
    \end{minipage}
    
    \begin{minipage}{\textwidth}
        \centering
        \includegraphics[width=0.88\textwidth]{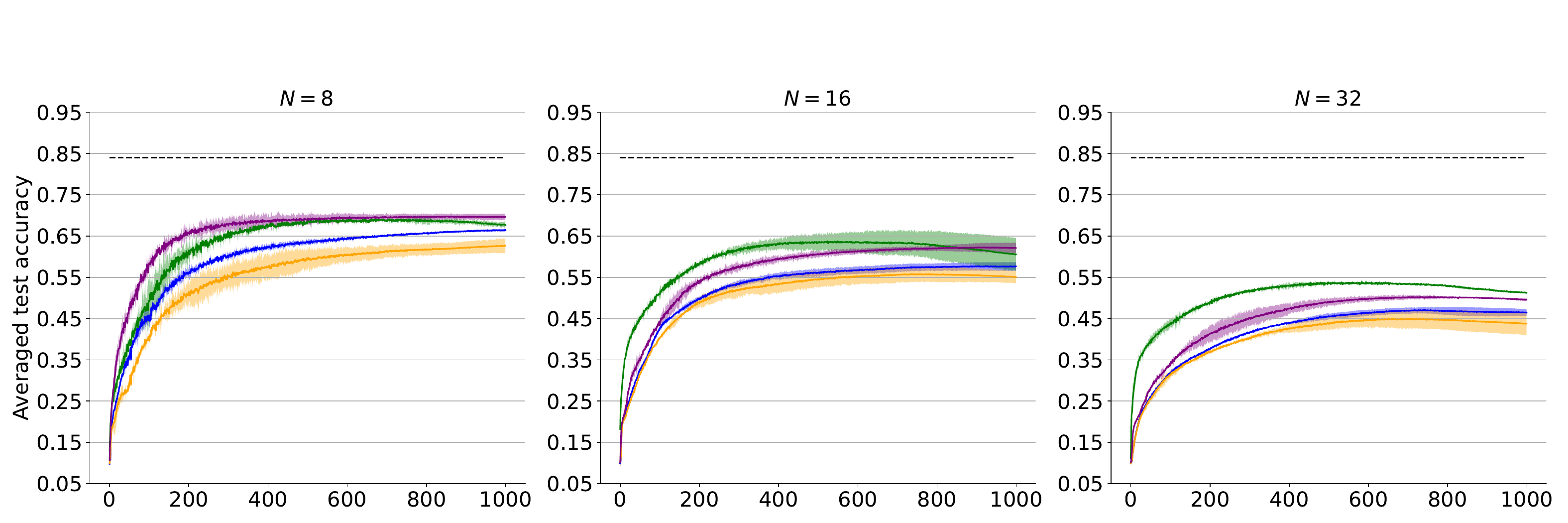}\\
        {\small (c) Chordal Topology}
    \end{minipage}
    
    \caption{Performance comparison of all approaches on the CIFAR10 dataset with ResNet20 architecture for various graph topologies, evaluated across different server counts ($N = 8, 16,$ and $32$). The y-axis represents the average test accuracy, computed over $3$ random seeds and averaged across all servers. The shaded regions indicate the $95\%$ confidence interval.}
     \label{fig:CIFAR10_test_acc}
\end{figure*}

\begin{itemize}
    \item \textbf{Centralized training} of neural networks serves as a critical baseline, providing a benchmark for performance evaluation. This approach involves training the model using all available data on a single, centralized system, ensuring the model benefits from the entire dataset without the constraints of data distribution across multiple servers. The results from centralized training offer a performance reference point to compare against distributed learning methods.
    \item \textbf{FedAvg} involves each (or a fraction of) server communicating its model parameters to a centralized server at every communication round. The central server aggregates these parameters and sends the updated model back to all servers. While FedAvg provides a centralized performance benchmark, its major drawback is the communication complexity, which scales linearly with the number of servers, potentially leading to significant communication overhead in large-scale deployments. In our implementation, we assume that \emph{all servers} send their model parameters to the central server at each communication round. Thus, our implementation resembles a centralized setting, but model updates at the edge servers occur locally.
    \item \textbf{Simple Weights} is a static consensus-based distributed learning method. In this approach, during the consensus step, each server assigns equal weights to its neighboring servers when aggregating model parameters. This method ensures uniform contribution from all neighboring nodes, providing a straightforward and balanced way to achieve consensus across the network.
    \item \textbf{Metropolis Weights} is similar to the Simple Weights approach but uses weights derived from the Metropolis-Hastings algorithm instead of assigning equal weights to neighboring servers during the consensus step. This method takes into account the degree of each server, ensuring a more balanced and efficient aggregation of model parameters based on network connectivity (see \eqref{eq:Metro}). 
\end{itemize}

\textbf{Note: Since the Centralized method has access to the whole dataset and the FedAvg method aggregates parameters from all the servers, it is expected that both will perform better compared to decentralized methods. They are used solely as centralized benchmarks.}

\subsection{Experimental Setup}

\subsubsection{Datasets} We present the analysis on three widely used computer vision datasets: \textit{MNIST}~\cite{lecun2010mnist}, \textit{CIFAR10}~\cite{krizhevsky2014cifar}, and \textit{CIFAR100}~\cite{krizhevsky2014cifar}. A detailed description of these datasets is provided in appendix \ref{sec:Dataset}.

\subsubsection{Network Topology} - Communication efficiency is a key factor in decentralized learning, and it depends on the average degree of the network's topology. For \textit{MNIST} and \textit{CIFAR10} datasets, we evaluate performance across ring, line, and chordal network topologies, considering different numbers of servers ($N = 8, 16, 32$). In the case of the \textit{CIFAR100} dataset, we conduct experiments using an undirected static exponential graph topology with $32$ servers. For more detail see appendix~\ref{sec:topologies}.

\subsubsection{Data Distribution}\label{subsubsec:data_distibution} - In cases of homogeneous data distribution across servers, \texttt{DYNAWEIGHT} behaves similarly to a simple weighting scheme. However, decentralized learning faces its primary challenge with heterogeneous, non-IID data. To address this, we test various approaches under heterogeneous data regimes. For all three classification datasets, servers with IDs that are multiples of 4 receive a uniform distribution across all classes, though with a limited number of data points per class. The remaining servers receive data from only 3 randomly chosen classes out of the 10 available for \textit{MNIST} and \textit{CIFAR10}, while servers not multiple of 4 for \textit{CIFAR100} are assigned data from 10 to 15 randomly selected classes out of the 100 available.

\subsubsection{Model Architecture and Hyperparameters}  
We employ well-established deep neural network architectures for classification tasks: \textit{LeNet}~\cite{lecun1998gradient} for \textit{MNIST}, \textit{ResNet-20}~\cite{he2016deep} for \textit{CIFAR10}, and \textit{ResNet-56}~\cite{he2016deep} for \textit{CIFAR100}. 
A detailed description of the model architectures and their corresponding hyperparameter configurations is  provided in Appendix~\ref{sec:model_and_hyperparameters}.

 \subsection{Results Discussion}
Figure~\ref{fig:MNIST_test_acc} illustrates the average test accuracy across epochs when training the \textit{LeNet} architecture on the \textit{MNIST} dataset, using ring, line, and chordal graph topologies with $N=8$, $16$, and $32$ servers. It is evident that \texttt{DYNAWEIGHT} achieves faster convergence compared to static weighting methods for all graph sizes and topologies. Notably, for $N=8$, all methods converge to $99\%$ test accuracy, which aligns with the performance of centralized training.\\

\noindent\textbf{Performance degradation with an increased number of servers} However, as the graph size increases, the performance gap between centralized training and the other methods becomes more pronounced. This occurs because, with larger graph sizes, each server receives fewer data points, leading to greater data heterogeneity. Consequently, the local gradient of each server becomes more misaligned with those of its neighboring servers. During the gossip step, this misalignment causes higher variance when aggregating model parameters from neighboring servers, which slows down learning and reduces the final test accuracy. Since \texttt{DYNAWEIGHT} dynamically adjusts the weights for neighboring servers based on their relative losses on local datasets, it outperforms static weighting methods for larger graph sizes ($N=16$ and $32$), achieving a $2-5\%$ improvement in converged test accuracy.

\begin{remark}
    When data heterogeneity is high across servers, decentralized learning methods struggle to match the test accuracy of centralized approaches. This is because decentralized models rely on local data, which may not fully represent the overall dataset, leading to imbalanced learning. As a result, even with adaptive schemes like \texttt{DYNAWEIGHT}, the performance tends to align more closely with that of FedAvg, which serves as the benchmark in such settings. 
\end{remark}

\noindent\textbf{Impact of Network Topology on Performance} Graph topologies with higher average degrees typically feature larger spectral gaps, which promote faster information exchange between nodes and consequently lead to quicker convergence in decentralized optimization tasks. This trend is clearly observed in our results, as shown in Figures~\ref{fig:MNIST_test_acc} and \ref{fig:CIFAR10_test_acc}. Specifically, the chordal graph achieves superior average test accuracy compared to both the ring and line graphs due to its higher average degree.\\

\begin{figure}[ht]
    \centering
    \includegraphics[width=0.88\linewidth]{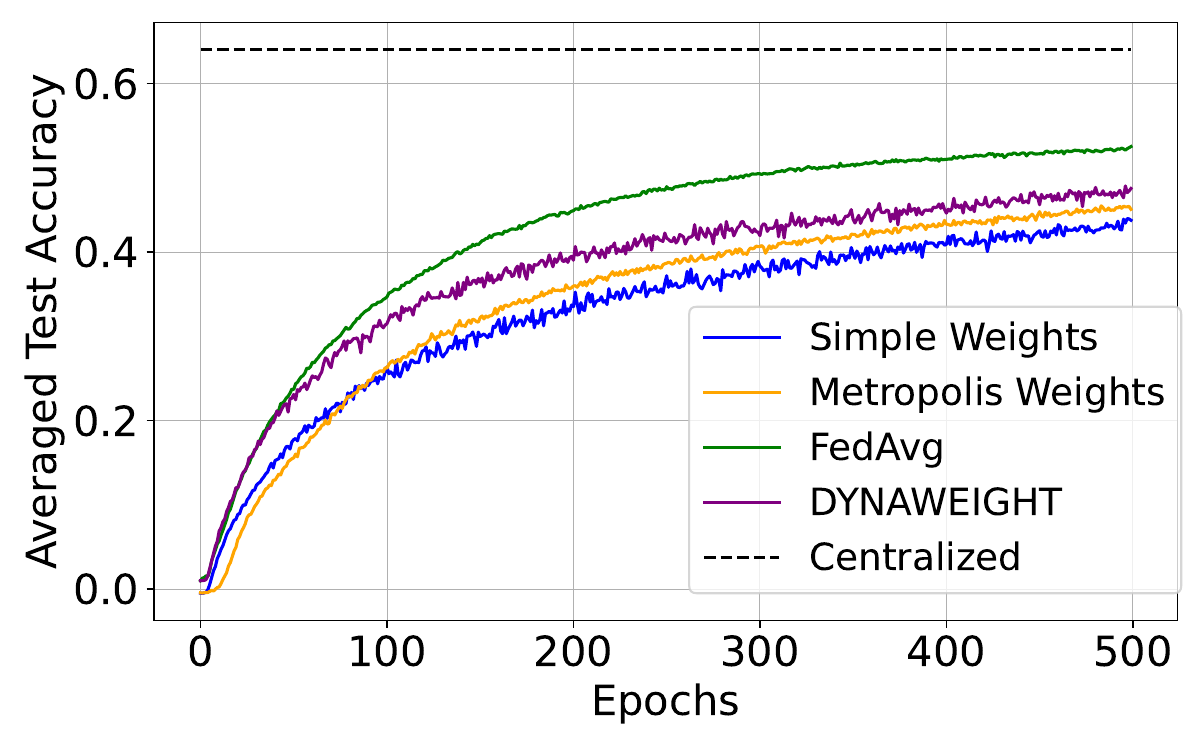}
    \caption{Performance comparison of all approaches on the CIFAR100 dataset with ResNet56 architecture for undirected static exponential graph with $32$ servers. The y-axis represents the average test accuracy across all servers.}
    \label{fig:cifar100_test_Acc}
\end{figure}

\noindent\textbf{Generalizability and Scalability} To assess the generalizability and scalability of \texttt{DYNAWEIGHT}, we evaluate its performance on the \textit{CIFAR10} dataset with the \textit{ResNet-20} architecture, and on the \textit{CIFAR100} dataset with the \textit{ResNet-56} architecture. Figure~\ref{fig:CIFAR10_test_acc} shows the average test accuracy of \textit{ResNet-20} trained on \textit{CIFAR10} across ring, line, and chordal topologies with $N=8$, $16$, and $32$ servers. As demonstrated, \texttt{DYNAWEIGHT} consistently converges faster than static weighting methods across all graph sizes and topologies. Notably, for graph sizes $N=8$ and $16$, \texttt{DYNAWEIGHT} outperforms static weighting methods with a significant performance gain of $8-10\%$. Even with $N=32$, \texttt{DYNAWEIGHT} retains a notable advantage, delivering an approximate $5\%$ improvement over static weighting approaches. Similarly, Figure~\ref{fig:cifar100_test_Acc} presents the average test accuracy of \textit{ResNet-56} trained on \textit{CIFAR100} with $N=32$ servers. Note that while \textit{MNIST} and \textit{CIFAR10} have fewer classes, \textit{CIFAR100} has 100 classes, and our experimental setup introduces significant data heterogeneity across servers. To accelerate decentralized learning, we adopt an undirected static exponential graph topology~\cite{10.5555/3540261.3541332}, which has been proven to balance the trade-off between communication and computation more effectively than other topologies. Once again, \texttt{DYNAWEIGHT} demonstrates faster convergence and achieves $2\%$ improvement in accuracy over static weighting methods.

\begin{figure}[ht]
    \centering
    \begin{tabular}{c}  
        \includegraphics[width=0.88\linewidth]{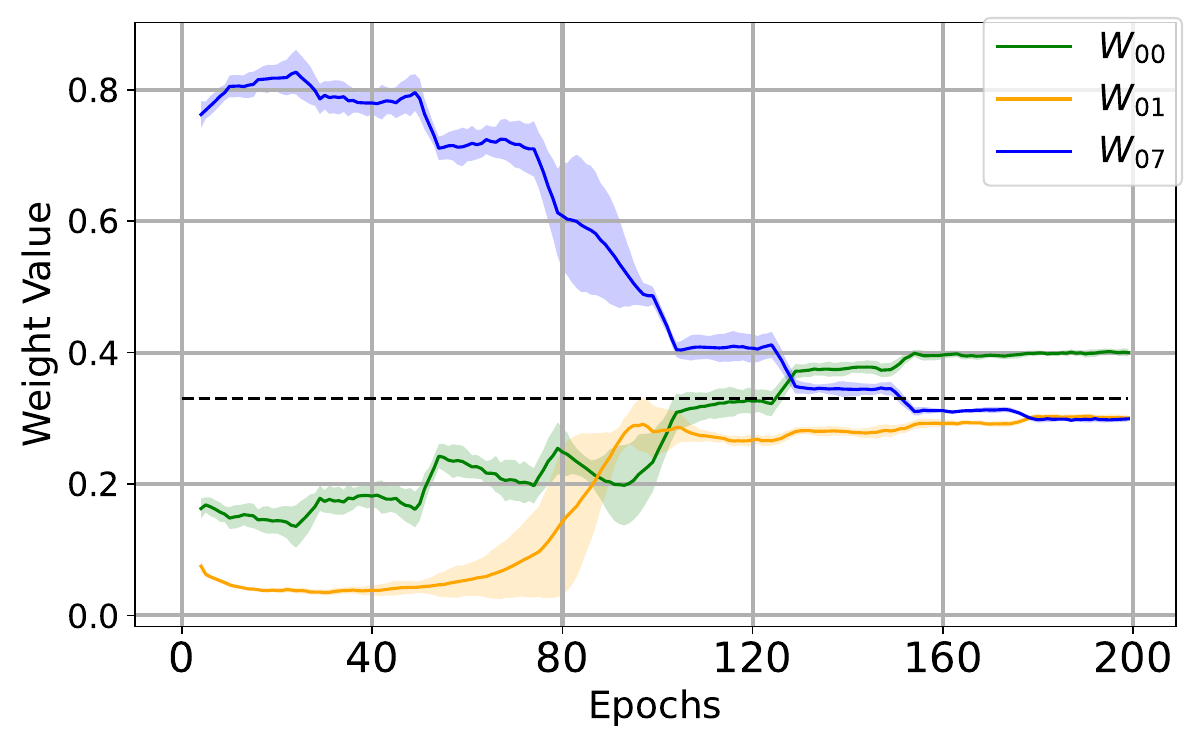} \\
        \includegraphics[width=0.88\linewidth]{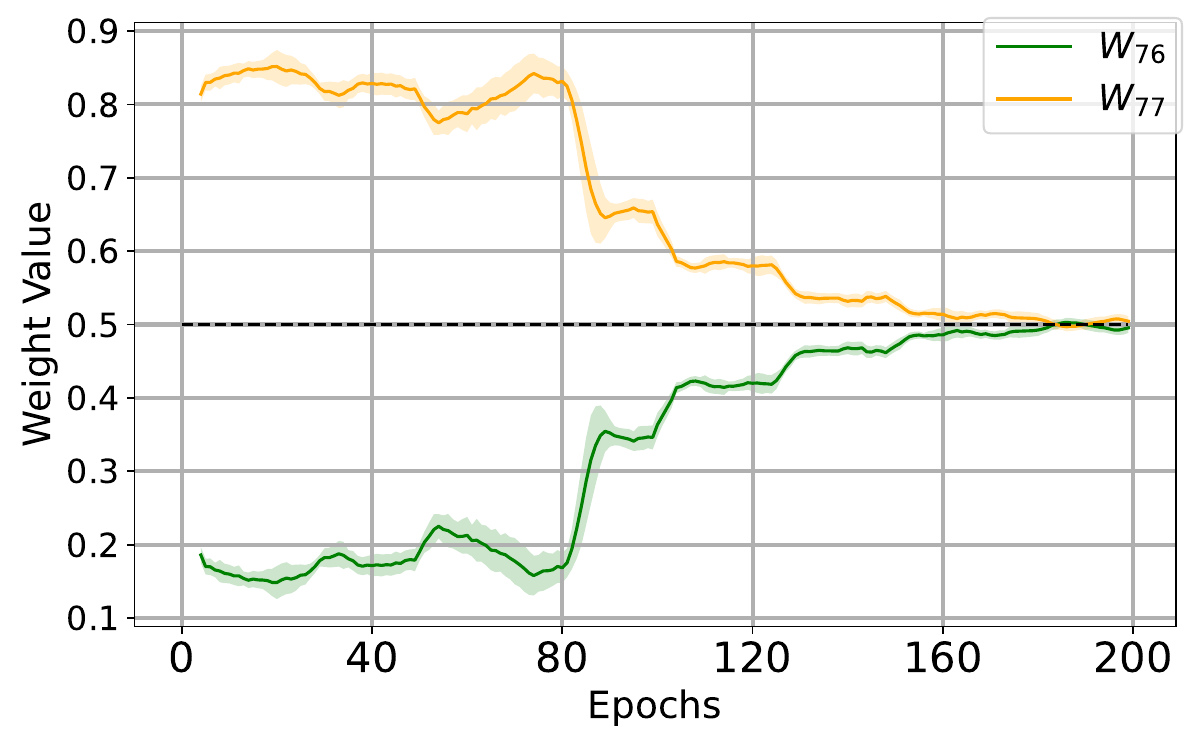}
    \end{tabular}
    \caption{Evolution of \texttt{DYNAWEIGHT} weights ($w_{ij}$) for the neighboring servers (including itself) of server 0 (top plot) and server 7 (bottom plot) in ring and line graph topologies, respectively, with a total of 8 servers, on MNIST dataset. The shaded regions indicate the $95\%$ confidence interval.}
    \label{fig:weight_evolution_plot}
\end{figure}

\noindent\textbf{Graph Weight Evolution} Figure~\ref{fig:weight_evolution_plot} illustrates the evolution of \texttt{DYNAWEIGHT} weights ($w_{ij}$) for server $0$ in the ring topology (top plot) and for server $7$ in the line topology (bottom plot), with a total of $8$ servers in the graph topology, on the MNIST dataset. In the ring topology, the neighboring nodes for server $0$ are servers $1$, $7$, and $0$ (itself), all of which contribute during the model-parameter aggregation (gossip step). The figure shows how the weight assigned to the model parameters of servers $0$, $1$, and $7$ evolves over the epochs during the aggregation process for server $0$. As explained in Section~\ref{subsubsec:data_distibution}, server $7$ has a more balanced data distribution across all classes, while servers $0$ and $1$ only have data from three classes. This data imbalance leads \texttt{DYNAWEIGHT} to initially assign a higher weight to server $7$. Over time, as the servers exchange information during the gossip steps, the weights for all neighboring servers converge to approximately 1/3, indicating that, after consensus, each server contributes equally to the aggregated model. A similar pattern is observed in the line topology. In this case, server $7$ has neighboring servers $6$ and $7$ (itself). Initially, \texttt{DYNAWEIGHT} assigns a higher weight to server $7$ due to its more balanced dataset, but as training progresses, the weights for both servers converge to 0.5, reflecting a more balanced contribution during the aggregation.\\

\begin{figure}[ht]
    \centering
    \includegraphics[width=0.88\linewidth]{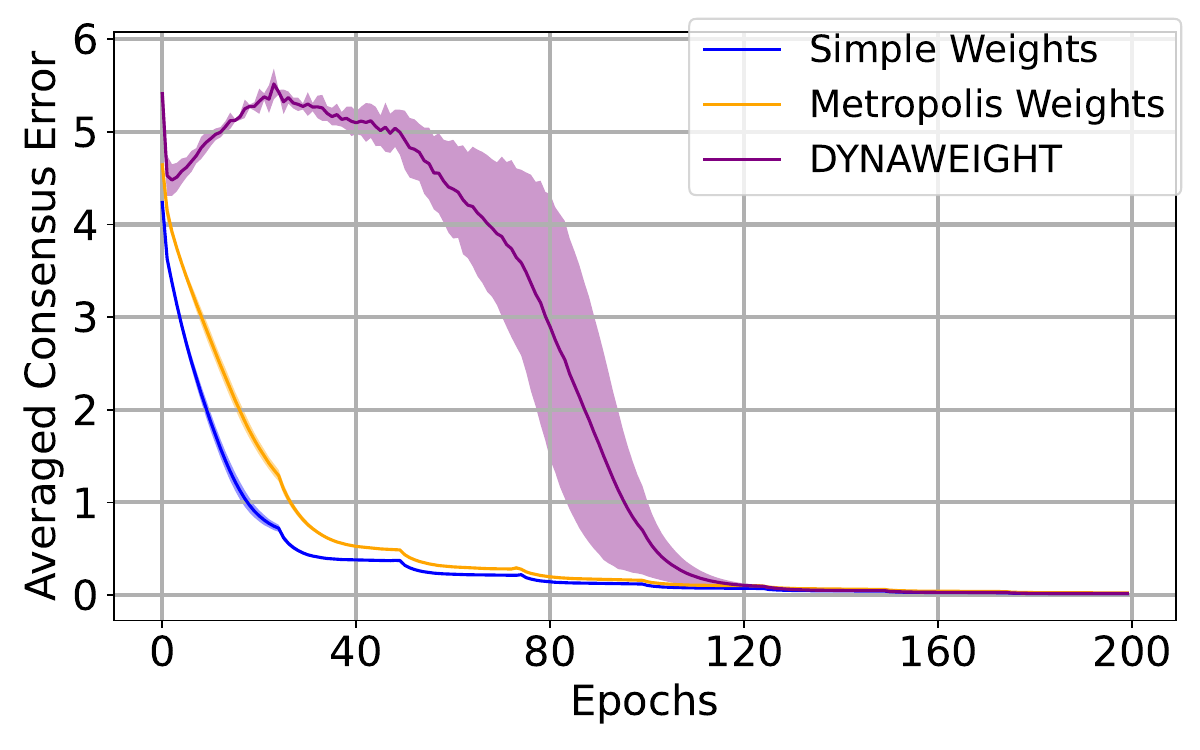}
    \caption{Consensus error on the MNIST dataset for the ring topology with 8 servers, averaged over $3$ random seeds and across all servers. The shaded regions indicate the $95\%$ confidence interval.}
    \label{fig:consensus_error_plot}
\end{figure}

\noindent\textbf{Consensus Error} Figure~\ref{fig:consensus_error_plot} shows the averaged consensus error on the MNIST dataset for a ring topology with $8$ servers. The consensus error is defined as the average sum of the Euclidean distances between each server's model parameters and the averaged model parameters across all servers. As shown in the figure, \texttt{DYNAWEIGHT}'s consensus error initially increases before gradually decreasing. In contrast, static weighting schemes cause the consensus error to decrease rapidly from the start, limiting the exploration of the parameter space by the server models. \texttt{DYNAWEIGHT} allows the consensus error to decrease more gradually, which facilitates more effective exploration of the parameter space by each server's model. This adaptive approach results in better convergence of test accuracy across various datasets, model architectures, graph topologies, and server sizes when compared to static weighting schemes.\\

\noindent\texttt{DYNAWEIGHT} provides a robust framework for decentralized learning, ensuring that sensitive data stays local to each server. Unlike centralized approaches, where data is aggregated or shared at a central location, \texttt{DYNAWEIGHT} enables each server to perform local computations while only sharing model parameters during the aggregation step. While \texttt{DYNAWEIGHT} does introduce some communication overhead compared to static weighting schemes due to the need for servers to broadcast both model parameters and additional scalar values such as loss and centrality, this overhead remains minimal in terms of computational and memory costs. These minor overheads are far outweighed by the advantages of adaptive weighting, which results in faster convergence and improved accuracy, especially in scenarios with heterogeneous data distributions.

While the framework is robust, addressing adversarial agents remains an area for future exploration. One approach could involve detecting and excluding servers with loss values that significantly deviate from the expected distribution, thereby reducing the impact of malicious or misbehaving agents. This adaptability underscores the framework's flexibility and potential for development into more secure, adversarial-resistant systems.

\section{Conclusion}\label{sec:Con}
In this paper, we introduced \texttt{DYNAWEIGHT}, an adaptive weighting framework designed to tackle the challenges of data heterogeneity in decentralized learning environments. Unlike static weighting schemes that rely solely on network connectivity, \texttt{DYNAWEIGHT} dynamically adjusts weights based on model performance on neighboring servers' datasets, ensuring more efficient and balanced learning. Our empirical results on \textit{MNIST}, \textit{CIFAR10}, and \textit{CIFAR100} demonstrate that \texttt{DYNAWEIGHT} converges faster than traditional non-adaptive schemes while incurring minimal computational and memory overhead. This framework enables more robust and scalable decentralized learning systems capable of handling diverse and unevenly distributed data efficiently. Due to resource constraints, we could not test on larger datasets like ImageNet. Future work will include a broader set of results and a theoretical analysis of \texttt{DYNAWEIGHT}.


\appendix
\section{Datasets}\label{sec:Dataset}
\begin{itemize}
    \item The \textit{MNIST}dataset~\cite{lecun2010mnist} is a widely used benchmark for evaluating image classification algorithms. It consists of $70k$ grayscale images of handwritten digits from $0$ to $9$, each sized $28X28$ pixels. The dataset is divided into $60k$ training images and $10k$ testing images.
    \item The \textit{CIFAR10}~\cite{krizhevsky2014cifar} image classification dataset consisting of 10 classes, comprising $60k$ color images, each sized $32X32$ pixels. Each class has $6000$ images, with $5000$ for training and $1000$ for testing.
    \item The \textit{CIFAR100}~\cite{krizhevsky2014cifar} dataset is a more complex image classification dataset, comprising $60k$ color images, each sized $32X32$ pixels, categorized into $100$ different classes. Each class has $600$ images, with $500$ for training and $100$ for testing.
\end{itemize}

\section{Network Topology}\label{sec:topologies}
\begin{itemize}
    \item Ring Graph - A ring graph is an undirected graph in which each node connects to its two immediate neighbors, forming a closed loop.
    \item Line Graph - A line graph is an undirected graph in which each node (except endpoints) connects to two neighbors.
    \item Chordal Graph - A chordal graph is an undirected graph in which every cycle of four or more vertices has a chord—an edge that connects two non-adjacent vertices in the cycle.
    \item Static Exponential Graph - A static exponential graph is a directed graph in which each node connects to exponentially increasing distances nodes. But in our experiments we consider it as an undirected graph.
\end{itemize}
\begin{figure}[ht]
    \centering
    \includegraphics[width=0.8\linewidth]{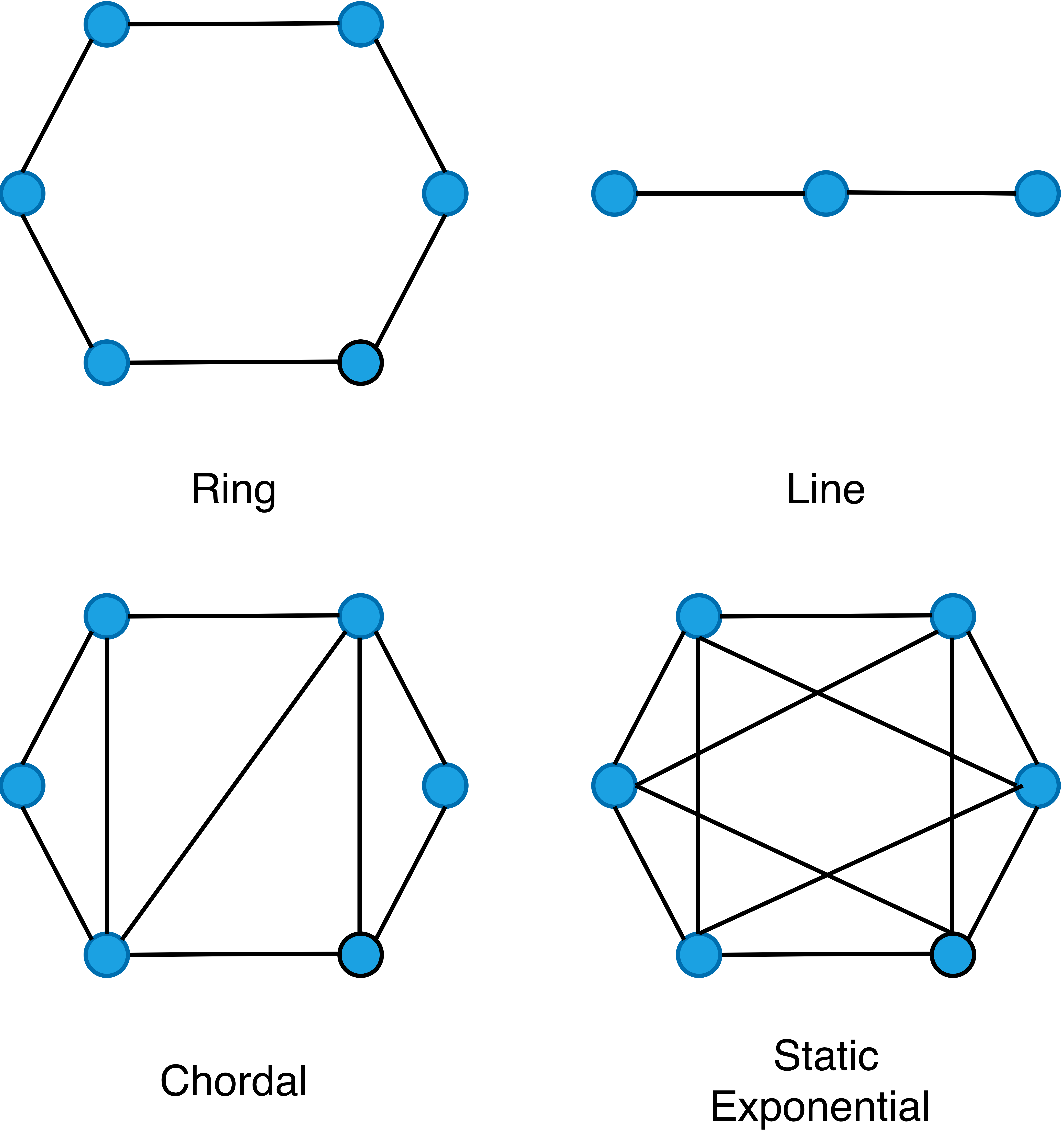}
    \caption{Ring Graph, Line Graph, Chordal Graph, and Static-Exponential Graph (undirected)}
    \label{fig:enter-label}
\end{figure}

\section{Model Architecture and Hyper-parameters}\label{sec:model_and_hyperparameters}
\begin{itemize}
    \item \textit{LeNet} -  For MNIST dataset classification, we use LeNet~\cite{lecun1998gradient} network architecture consisting of two convolutional layers followed by two fully-connected layers. The first convolutional layer has 32 filters with a kernel size of 3x3, and the second has 64 filters with the same kernel size. Each convolutional layer is followed by a max-pooling layer with a filter size of 2x2 and a stride of 2. After the convolutional layers, there are two fully-connected layers: the first has 128 units, and the second outputs 10 units, corresponding to the 10 classes in the dataset. We use the same model architecture for all baselines and servers, along with consistent hyperparameters. The batch size is set to 16, and we use the Adam optimizer with an initial learning rate of $10^{-4}$, which changes over the training epochs. For $N=8$ and $N=16$ servers, the learning rate is halved every 20 epochs, and the number of consensus steps are set to $C=1$. For $N=32$ servers, the learning rate remains constant until the $100^{th}$ epoch, after which it is exponentially reduced to $10^{-6}$ by the end of training, and the number of consensus steps are set to $C=2$.
    \item \textit{ResNet-20} - For CIFAR10 dataset classification, we use the standard ResNet-20~\cite{he2016deep} architecture with $0.27M$ trainable parameters. We use the same model architecture for all baselines and servers, along with consistent hyperparameters. The batch size is set to 16, and we use the Adam optimizer with an initial learning rate of $10^{-4}$, which changes over the training epochs. For all server setting (i.e. $N = 8, 16,$ and $32$ ), the learning rate remains constant until the $100^{th}$ epoch, after which it is exponentially reduced to $10^{-6}$ by the end of training, and the number of consensus steps are set to $C=2$.
    \item \textit{ResNet-56} - For CIFAR100 dataset classification, we use the standard ResNet-56~\cite{he2016deep} architecture with $0.85M$ trainable parameters. We use the same model architecture for all baselines and servers, along with consistent hyperparameters. The batch size is set to 16, and we use the Adam optimizer with an initial learning rate of $10^{-4}$, which we decay linearly over the training epoch, and the number of consensus steps are set to $C=2$.
\end{itemize}

\section{GenAI Usage Disclosure}
We used OpenAI's ChatGPT (GPT-4) to assist with minor paraphrasing and language refinement during the editing process. All technical content and original writing are our own, and no text was generated by the tool beyond rewording.
\bibliographystyle{ACM-Reference-Format}
\balance
\bibliography{sample-base}

\end{document}